# Learning Parameters for a Generalized Vidale-Wolfe Response Model with Flexible Ad Elasticity and Word-of-Mouth

Yanwu Yang[1], Baozhu Feng[1], Daniel Zeng[2,3]

[1]School of Management, Huazhong University of Science and Technology, Wuhan, China
[2]Department of Management Information Systems, University of Arizona, USA
[3]Institute of Automation, Chinese Academy of Sciences, Beijing, China

**Abstract**: In this research, we investigate a generalized form of Vidale-Wolfe (GVW) model. One key element of our modeling work is that the GVW model contains two useful indexes representing advertiser's elasticity and the word-of-mouth (WoM) effect, respectively. Moreover, we discuss some desirable properties of the GVW model, and present a deep neural network (DNN)-based estimation method to learn its parameters. Furthermore, based on three realworld datasets, we conduct computational experiments to validate the GVW model and identified properties. In addition, we also discuss potential advantages of the GVW model over econometric models. The research outcome shows that both the ad elasticity index and the WoM index have significant influences on advertising responses, and the GVW model has potential advantages over econometric models of advertising, in terms of several interesting phenomena drawn from practical advertising situations. The GVW model and its deep learning-based estimation method provide a basis to support big data-driven advertising analytics and decision makings; in the meanwhile, identified properties and experimental findings of this research illuminate critical managerial insights for advertisers in various advertising forms.

# 1. Introduction

Advertising is an important component in the marketing mix for a firm, which includes a variety of promotional ways across a rich set of media vehicles. Advertising response is one of kernel concepts in the advertising field, which relates market responses (e.g., sales or market share) to advertising expenditure. Essentially, the ad-response relationship captures the potential influence of advertising expenditure and related market determinants on market responses of interest. More importantly, it's the basis for various advertising decisions such as budget planning and media selection. Thus, for decades, advertising responses have been a research hotspot attracting plenty of attentions from both academia and industries (e.g., [1,2]).

In the literature, quite a few advertising response models have been developed to quantitatively describe the ad-response relationship. According to the mathematic form, advertising response models can be categorized into two classes: analytic models and econometric models. There was a research surge on advertising response models from 1970s to 1980s. In the last two decades, the Internet has witnessed the advent of a large number of digital media vehicles (e.g., search portals, social media platforms, e-commerce platforms, online gaming, mobile apps, online videos, banners, etc.) that promised a variety of novel advertising forms[3]. In the meanwhile, advertising environments have evolved into a vastly complex communication system[4]. This calls for another revisiting of advertising response models.

The objective of this study is to investigate a generalized form of Vidale-Wolfe advertising model (GVW) and its applicability. Our motivation for choosing VW-type models lays on the fact that they combine several important market factors that influence advertising decisions, such as the carryover effect of past advertising on current sales, the saturation level, the possible diminishing returns to cumulative advertising expenditure, and the word-of-mouth (WoM) effect, in a time-varying manner[5]. In this research, we present a generalized form of VW model, which is dynamic in nature and flexible in terms of parameters. First, one key element of our modeling work is that the GVW model explicitly represents an advertiser's elasticity with respect to advertising expenditure and the WoM effect among potential consumers by two additional indexes, respectively. As observed by prior studies, the advertising elasticity[6] and the word-of-mouth effect[7,8] are highly prominent in online advertising, especially in e-commerce markets. Second, we discuss desirable properties of the GVW model, and develop a deep neural network (DNN)-based method to learn its modeling parameters. Furthermore, based on three real-world datasets, including a benchmark dataset on traditional advertising and two datasets obtained from advertising campaigns by two large e-commerce companies on Google AdWords and Facebook Ads, respectively, we conduct computational experiments to validate our model and identified properties. In addition, we also discuss potential advantages of the GVW model over econometric models.

On one hand, methodologically, the GVW model and its deep learning-based estimation method provide a basis to support big data-driven advertising analytics and decision makings. On the other hand, identified properties and experimental findings of this research illuminate critical managerial insights for advertisers in various advertising forms.

The remainder of this paper is organized as follows. The next section provides a review of the VW model. Section 3 presents a generalized VW model. Section 4 discusses some desirable properties analytically. Section 5 presents a deep neural network (DNN)-based estimation method. Section 6 reports experimental results to validate our model and identified properties, and discusses potential advantages of the GVW model over econometric models. Finally, we conclude this work in Section 7.

## 2. Review of the Vidale-Wolfe Model

Vidale and Wolfe[9] developed a differential equation to capture the advertising response dynamics, which is given as follows.

$$\dot{x} = \rho u(1-x) - \delta x, \ x(0) = x_0 \qquad (1)$$

In Equation (1), i.e., the VW model, $x$ and $1-x$ represent the sold portion and unsold portion of the potential market (i.e., a fixable pool of customers), respectively; $u$ represents a firm's advertising effort at time t. The pioneering VW model has been developed by introducing a concept called advertising effectiveness (i.e., ρ) that describes sales (or market share) generated from each advertising dollar at the level of zero sales. Specifically, ρ denotes the ad effectiveness index that represents the effectiveness of advertising expenditure, i.e., the response to advertising that acts positively on the unsold market share. In addition, δ denotes the decay index describing the loss of customers probably due to forgetting and competition that acts negatively on the sold portion of the market. In summary, the VW model encapsulates the two opposite forces on the ad-response relationship by using two parameters, namely ρ and δ, in the positive and negative direction, respectively.

Another related influential advertising response function is the Lanchester model[10], which is given as

$$\dot{x}_i = \rho_i u_i (1 - x_i) - \rho_j u_j x_i, \ \{i,j\} = \{1,2\}. \qquad (2)$$

where $u_i$ and $u_j$ denote the advertising efforts of advertiser i and j, respectively. Note that Equation (2) is specified for the duopolistic competition where $x_i + x_j \leq 1$.

As noticed by Little[11], the Lanchester model can be interpreted as a competitive generalization of VW model. A large body of contemporary research in advertising has extended the VW and Lanchester models to various situations. In the following section, we analyze the evolution of VW and Lanchester models in an integrative way, rather than introducing them separately, because these two models formally share a set of common parameters.

## 3. The GVW Model

In this section, we present a generalized form of the VW model (i.e., the GVW model). In detail, we will first examine two important components in the advertising effectiveness: the advertising effort and its potential Word-of-Mouth effect, through a systematic analysis of the extant literature on advertising response models. Then we present the GVW model which is developed by extending the VW model with the two additional indexes.

### 3.1 The Advertising Effort and Elasticity

In the VW and Lanchester models, the linear form of advertising effort ($u$) assumes a uniform return from each advertising dollar, which is neither realistic nor attainable through advertising practices. Theoretically, the function form of advertising effort, i.e., $u = f(b)$ where b represents advertising expenditure, should be in line with the law of diminishing marginal utility (i.e., the marginal return decreases as the investment increases, holding other factors constant) in order to render these models more applicable.

In this stream, plentiful research (e.g., [12]) investigated desirable forms of advertising effort. Either in the VW model under the monopolistic setting or in the Lanchester model under the competitive setting, the exponential form ($b^\alpha$), as suggested by Little[11], has been widely accepted because it exhibits the trend of diminishing marginal returns in advertising activities in a quite simple way[13]. In the exponential form, $\alpha$ stands for the ad elasticity, which we will explore in detail in Section 3.3. In particular, analytically, the exponential form of advertising effort makes the pulsing policy superior over the uniform policy.

The ad elasticity index is usually fixed as $1/2$ (i.e., $u = b^{1/2}$), which raises a favorable property for response functions, based on which analytical solutions can be derived[14,3]. However, Erickson[15] empirically estimated the ad elasticity index (i.e., the power index of the exponential form), found its value ranges from 0.116 to 0.726. Thus, it's of necessity to take the ad elasticity as a variable, rather than a constant.

## 3.2 The Word-of-Mouth Effect

In order to capture the additional effect generated from advertising efforts (i.e., potential communications between individuals comprising the sold portion and those comprising the unsold portion of the potential market), Sethi[16] extended the VW model by using a square-root form of the untapped potential (i.e., $\sqrt{1-x}$, where x denotes the sold portion) to capture the possible Word-of-Mouth (WoM) effect. Sethi's work is coincident with Sorger's modification[17] on the Case model[18]. Mathematically, $\sqrt{1-x}$ can be approximated by $1 - x + x(1-x)$. The positive effect raised by the modification from the linear form in the VW and Lanchester models to the nonlinear form is called an excess advertising term (i.e., $x(1-x)$).

Likewise, the square-root form of the unsold market makes the analytical analysis possible in certain settings. However, as far as we knew in the extant literature, there is little empirical evidence for the square-root form of the WoM Effect.

## 3.3 A Generalized Form of the Vidale-Wolfe Model

In order to fit flexible decision scenarios in various advertising forms, we present a generalized form of VW model (GVW), given as follows.

$$\dot{x} = \rho b^{\alpha}(1-x)^{\beta} - \delta x, x(0) = x_0 \qquad (3)$$

In the GVW model, i.e., Equation (3), $\alpha$ is the ad elasticity index is represented as the percentage change in the advertising effort to the change of one percentage in the advertising expenditure, i.e., $(\Delta b^{\alpha}/b^{\alpha})/(\Delta b/b)$, which can be interpreted as the effective advertising effort elasticity connecting the change in effective advertising effort to the change in advertising expenditure; $\beta$ is the WoM (i.e., word-of-mouth) index representing an additional process of WoM communication between the sold portion and the unsold portion of the potential market. Mathematically, $(1-x)^{\beta}$ can be approximated by $1 - x + 2(1-\beta)x(1-x)$, where $(1-\beta)$ quantifies the communication level between the sold and unsold portions. The value of $\beta$ might depend on application domains. For example, the reviewing feature of e-commerce entitles consumers to interact with each other, which thus raises a high WoM Effect. Specifically, the WoM Effect (i.e., $(1-\beta)$) increases as $\beta$ decreases. We will explore it in detail later in Sections 4 and 6.3.

## 4. Properties

In this section, we study some desirable properties of the GVW model. For more intuitive understanding, we discuss the response to a rectangular pulse of advertising and the steady-state response.

As for a rectangular pulse of advertising, we mean that a constant advertising, $b(t) \equiv b_0$, which is started at $t = 0$ and lasts until $t = T$ when it drops to zero. For $t \in [0, T]$, we have

$$\dot{x} = \rho b_0^{\alpha}(1-x)^{\beta} - \delta x. \quad (4)$$

Furthermore, we expand the nonlinear expression $(1-x)^{\beta}$ into Taylor series in $x = 0$ maintaining the second order, which is given as

$$(1-x)^{\beta} \approx 1 - \beta x + \frac{\beta(\beta-1)}{2}x^2. \quad (5)$$

Substituting Equation (5) into (4) yields

$$\dot{x} = k_1 x^2 + k_2 x + k_3, \quad (6)$$

where $k_1 = \frac{\rho\beta(\beta-1)b_0^{\alpha}}{2} \leq 0$, $k_2 = -\rho\beta b_0^{\alpha} - \delta < 0$, $k_3 = \rho b_0^{\alpha}$.

Solving $0 = k_1 x^2 + k_2 x + k_3$ yields

$$\hat{x} = \frac{-k_2 \pm \sqrt{k_2^2 - 4k_1 k_3}}{2k_1}.$$

Let $x = z + \hat{x}$, then

$$\dot{z} = (2k_1\hat{x} + k_2)z + k_1 z^2. \quad (7)$$

Solve Bernoulli's Equation (7), we obtain

$$z = [\frac{k_1}{2k_1\hat{x}+k_2}(e^{-(2k_1\hat{x}+k_2)t} - 1) + \frac{1}{x_0-\hat{x}}e^{-(2k_1\hat{x}+k_2)t}]^{-1}.$$

Thus, we have

$$x(t) = \begin{cases} [\frac{k_1}{2k_1\hat{x}+k_2}(e^{-(2k_1\hat{x}+k_2)t} - 1) + \frac{1}{x_0-\hat{x}}e^{-(2k_1\hat{x}+k_2)t}]^{-1} + \hat{x}, & 0 \leq t \leq T \\ x(T)e^{-\delta(t-T)}, & t > T \end{cases}. \quad (8)$$

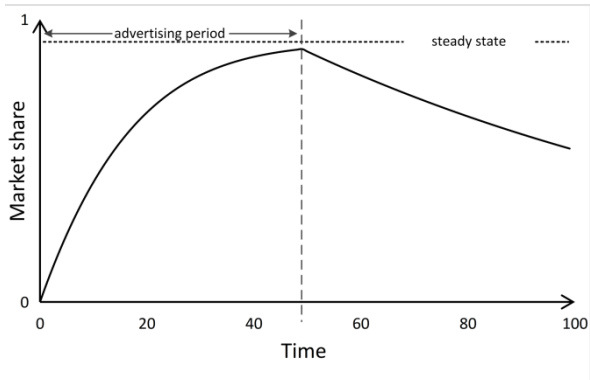

Figure 1. The Response to a Rectangular Pulse of Advertising (the GVW Model)

The evolution of market share described in Equation (8) is illustrated in Figure 1. Note that the steady-state level of market share ($\bar{x}$) corresponds to a constant advertising level ($\bar{b} = b_0$).

Basically, the evolution pattern of market share over time defined by the GVW model is consistent with that by the VW model. In the GVW model, the increasing period of market share is determined by all four modeling parameters, and the decreasing period is affected by the ad decay index ($\delta$). We will study the effects of the ad elasticity index ($\alpha$) and the WoM index ($\beta$) on market share in the increasing period in Section 6.3.

Next, we discuss the steady-state response of market share. In the steady state, the market share remains unchanged. That is, $\dot{x} = \rho \bar{b}^\alpha (1-x)^\beta - \delta \bar{x} = 0$.

Then we can obtain the advertising level in the steady state, which is given as

$$\bar{b} = [\frac{\delta \bar{x}}{\rho(1-\bar{x})^\beta}]^{\frac{1}{\alpha}} \qquad (9)$$

Based on Equation (9), we can obtain the relationship between steady-state market share and advertising, which is illustrated in Figure 2 (left). From Figure 2 (left), we can see that, as expected, the steady-state market share is positively related to the steady-state advertising level. Moreover, it exhibits the trend of diminishing marginal returns.

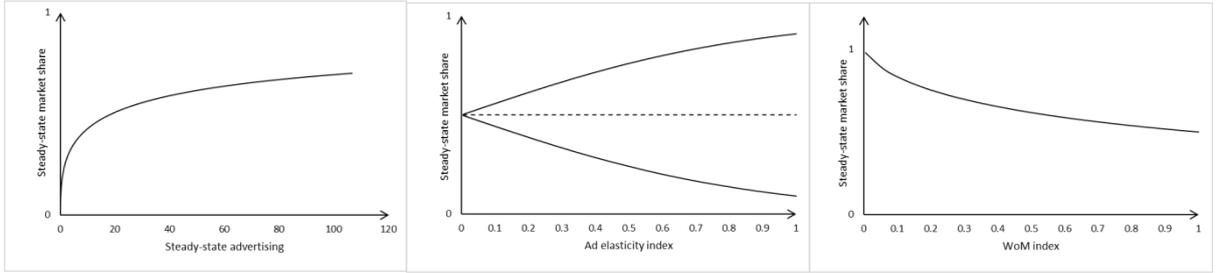

Figure 2. The Steady-State Response (the GVW Model)

Next, we examine the effect of the ad elasticity on the steady-state market share. The derivative of the steady-state ($\bar{x}$) with respect to the ad elasticity index is $\frac{d\bar{x}}{d\alpha} = \left(\frac{\partial \bar{x}}{\partial \bar{b}}\right)\left(\frac{\partial \bar{b}}{\partial \alpha}\right) + \frac{\partial \bar{x}}{\partial \alpha}$. From Equation (9), we can obtain

$$\begin{cases} \frac{\partial \bar{x}}{\partial \alpha} < 0, \bar{x} < \tilde{x} \\ \frac{\partial \bar{x}}{\partial \alpha} > 0, \bar{x} > \tilde{x} \end{cases}, \qquad (10)$$

where $\tilde{x}$ satisfies $\frac{\delta \tilde{x}}{\rho(1-\tilde{x})^\beta} = 1$.

Equation (10) characterizes the relationship between the steady-state market share and the ad elasticity. This indicates that, the effect of the ad elasticity on the steady–state market share depends on whether the steady state market share is above or below a threshold value, as illustrated in Figure 2 (middle).

Then we discuss the effect of the WoM index on the steady-state market share. The derivative of the steady-state with respect to the WoM index is $\frac{d\bar{x}}{d\beta} = \left(\frac{\partial \bar{x}}{\partial \bar{b}}\right)\left(\frac{\partial \bar{b}}{\partial \beta}\right) + \frac{\partial \bar{x}}{\partial \beta}$. From Equation (9), we can obtain $\frac{\partial \bar{x}}{\partial \beta} < 0$. As illustrated in Figure 2 (right), the steady-state market share monotonically increases with the WoM effect (i.e., $(1 - \beta)$).

## 5. A Deep Neural Network (DNN)-based Parameter Estimation Method

In this section we develop a Deep Neural Network (DNN)-based estimation method to assess the parameters of the GVW model (i.e., Equation 3). The DNN is particularly effective for estimating nonlinear models on large and noisy datasets[19].

The estimation problem of the GVW model can be formulated as follows.

$$\begin{cases} \min_{\Theta}\{\varepsilon_1 = \sum_{l=1}^{T}(Y_{t_l} - x_{t_l})^2\} \\ \dot{x} = \rho b^{\alpha}(1-x)^{\beta} - \delta x, x(0) = x_0 \end{cases}, \quad (11)$$

where $Y_{t_l}$ is the observed market share at time $t_l$, $\varepsilon_1$ is the error between observed values and model predictions, $\Theta(\rho, \alpha, \beta, \delta)$ is the vector of parameters to be estimated. We aim to obtain a set of estimates of model parameters by minimizing the error $\varepsilon_1$.

However, it is impossible to avoid integration of the nonlinear dynamic equation when solving problem (11). To this end, we turn to solve the following problem which is equivalent to problem (11).

$$\min_{\Theta}\{\varepsilon_2 = \sum_{l=1}^{T}(\frac{dy}{dt}\Big|_{t=t_l} - \frac{dx}{dt}\Big|_{t=t_l})^2\}, \quad (12)$$

where y satisfies $y(t) = Y_t$.

In order to solve problem (12), we need to calculate the value of $\frac{dy}{dt}\Big|_t$. In this research, we construct a DNN to approximate $\frac{dy}{dt}\Big|_t$. The DNN consists of one input layer with units $Z = (z_1, z_2, \cdots z_{K_0})$, m (m > 1) hidden layers with output activations $H_i = (h_1^i, h_2^i, \cdots h_{K_i}^i)$ (i = 1, $\cdots$ m), and one output layer with units $\bar{Y} = (\bar{y}_1, \bar{y}_2, \cdots \bar{y}_K)$. Output activations for hidden layers (i = 1, $\cdots$ m) and units for the output layer (i = m + 1) can be calculated as follows.

$$h_k^i = \sigma^i(\bar{h}_k^i), \quad \bar{h}_k^i = \sum_{j=1}^{K_{i-1}} w_{kj}^i h_j^{i-1} + \bar{b}_k^i, \quad k = 1, \cdots K_{i-1}, \quad (13)$$

where $w_{kj}^i$ and $\bar{b}_{kj}^i$ represent the constant weight and the bias term, respectively, $\sigma^i$ is the activation function, $K_i$ is the number of nodes in the $i-\text{th}$ hidden layer. Note that $h_k^0 = z_k$, $k = 1, \cdots K_0$ and $h_k^{m+1} = \bar{y}_k$, $k = 1, \cdots K$, where $K_0$ is the number of input units and $K$ is the number of output units.

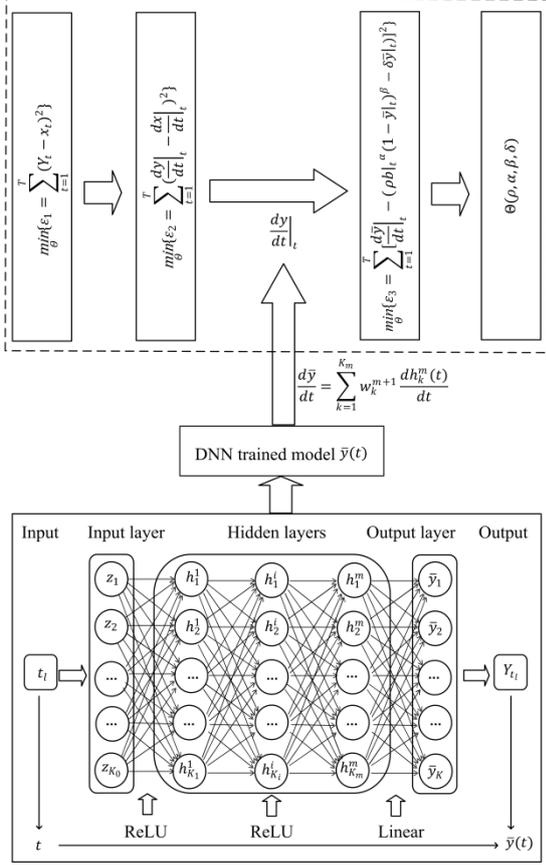

Figure 3. The Deep Neural Network (DNN) based Parameter Estimation Method for the GVW Model

The structure of the DNN model is shown in Figure 3. We set the input $Z = t_l$ and the output $\bar{Y} = Y_{t_l}$. Moreover, the activation function for a hidden layer is nonlinear, which is specified as the rectified linear unit (ReLU). The ReLu typically raises a better performance in the training of deep neural networks on complex datasets, and its simple derivative form provides a convenience for calculations[20]. Furthermore, we use the same $\sigma$ for hidden layers, i.e., $\sigma^i(z) = \sigma(z) = \begin{cases} z, z > 0 \\ 0, z \leq 0 \end{cases}$, and its derivative $\sigma'(z) = \begin{cases} 1, z > 0 \\ 0, z \leq 0 \end{cases}$. Additionally, the activation function for the output layer is linear, i.e., $\sigma^{m+1}(z) = z$.

The DNN is trained to obtain values of weights and biases that minimize the loss function $f(\bar{\Theta})$, where $\bar{\Theta}$ represents the unknown parameters of the DNN comprising the weights and biases. The loss function is represented as the sum of squares of the difference between the output $Y_{t_l}$ and the corresponding predicted value $\bar{y}(t_l)$ obtained by the DNN, i.e., $f(\overline{\Theta}) = \sum_{l=1}^{T}[Y_{t_l} - \bar{y}(t_l)]^2$. The training procedure discussed above is based on the Levenberg-Marquardt algorithm[21].

Then we use $\frac{d\bar{y}}{dt}$ to approximate $\frac{dy}{dt}$. The output of the DNN, i.e., $\bar{y}(t)$ is represented as follows.

$$\bar{y}(t) = \sum_{k=1}^{K_m} w_k^{m+1} h_k^m(t) + \bar{b}^{m+1} \qquad (14)$$

Differentiating (14) with respect to t, we obtain

$$\frac{d\bar{y}}{dt} = \sum_{k=1}^{K_m} w_k^{m+1} \frac{dh_k^m(t)}{dt}, \text{ and}$$

$$\frac{dh_k^i(t)}{dt} = \text{ReLU}'(\bar{h}_k^i) \sum_{j=1}^{K_{i-1}} w_{kl}^i \frac{dh_j^{i-1}(t)}{dt}, k = 1, \cdots K_i, i = 1, \cdots m. \qquad (15)$$

Given that the DNN parameters $\overline{\Theta}$ are determined, we obtain the following simpler optimization problem:

$$\min_{\Theta}\{\varepsilon_3 = \sum_{l=1}^{T}[\frac{d\bar{y}}{dt}\Big|_{t=t_l} - (\rho b|_{t=t_l}^{\alpha}(1 - \bar{y}|_{t=t_l})^{\beta} - \delta\bar{y}|_{t=t_l})]^2\}. \qquad (16)$$

Thus, the parameter estimation for the GVW model can be achieved by solving the nonlinear programming problem (Equation 16). The detailed procedure for DNN-based parameters estimation of the GVW model is described in Algorithm 1.

**Algorithm 1**. (DNN-based Parameters Estimation)

**Input:** the advertising expenditure $b_{t_l}$, the market share $Y_{t_l}$, $l \in (1, \cdots T)$

**Output:** the vector of model parameters $\Theta(\rho, \alpha, \beta, \delta)$

**Procedure**

Step 1: Formulate the original optimization problem (11), and transform problem (11) into problem (12).

Step 2: Train the DNN model (call Sub-Procedure DNN Training).

Step 3: Calculate $\frac{d\bar{y}}{dt}$ according to Equation (15), and use it to approximate $\frac{dy}{dt}$.

> Step 4: Transform problem (12) into problem (16).
>
> Step 5: Obtain parameter estimates for the GVW model by solving problem (16).
>
> **End Procedure**
>
> **Sub-Procedure: DNN Training**
>
> Step 1: Construct the DNN.
>
> Step 2: Train the DNN to obtain $\bar{\Theta}$.
>
> Step 3: Obtain $\bar{y}(t)$ according to Equation (14).
>
> **End Sub-Procedure**

## 6. Experimental Study

The nonlinearity feature of the GVW model with respect to advertising expenditure, raised by the two newly added parameters (i.e., the ad elasticity and the WoM indexes), is attractive in terms of its flexibility and adaptability to complex advertising situations. In the meanwhile, this feature makes the GVW untractable. In this section, we design computational experiments to validate the proposed GVW model and identified properties.

Our experimental evaluation serves the following twofold purposes. First, we are intended to apply the GVW model on three separate datasets about a traditional advertising, search advertising (on Google Adwords), and social media advertising (on Facebook Ads), respectively, in order to verify whether the two newly added modeling parameters, namely the ad elasticity index ($\alpha$) and the WoM index ($\beta$), work in practical situations. Second, we conduct sensitivity analysis to further explore properties of our GVW model. Specifically, we evaluate possible influences of two newly added indexes on the relationship between market share and advertising expenditure in our GVW model. Next, we provide details about our experimental setup and some key results.

### 6.1 Data Description

We collected three realworld datasets from advertising campaigns. The first dataset is annual domestic sales and advertising by the Lydia Pinkham Medicine Company from 1907 to 1960, which has widely used in prior adverting research[22]. The second dataset records search advertising campaigns by a large U.S. electronic commerce retailer during a 33-month period, spanning 4 calendar years, which contains approximately 7 million records from almost 55,000 advertisements. The third dataset records historical information of social media advertising campaigns on Facebook by a European electronic commerce retailer operating in 8 countries. It contains 62,802 records from 95 advertising campaigns during a 20-month period. Finally, we also generate data from historical advertising reports to support computational experiments to verify properties of the GVW model.

**6.2 Empirical Analysis**

We estimate the GVW model on the three datasets using Algorithm 1. The DNN is trained to obtain an approximation of $\left.\frac{dy}{dt}\right|_t$ in our deep learning-based algorithm of parameter estimation, as described in Section 5. We first present the DNN training setting. The DNN models are with hidden layers size of $4 \times 8$, $16 \times 32$, $32 \times 32$ for Lydia Pinkham, Google and Facebook datasets, respectively. For the DNN training, key hyperparameters related to the combination coefficient in the Levenberg-Marquardt algorithm consists of the initial value (InV), an increase factor (InF) to increase its value, and a decrease factor (DeF) to decrease its value, which are set as InV=0.001, InF=10, DeF=0.1. The maximum number of epochs to train is 1000. In addition, the Mean Squared Error (MSE) is adopted to evaluate the performance of the DNN.

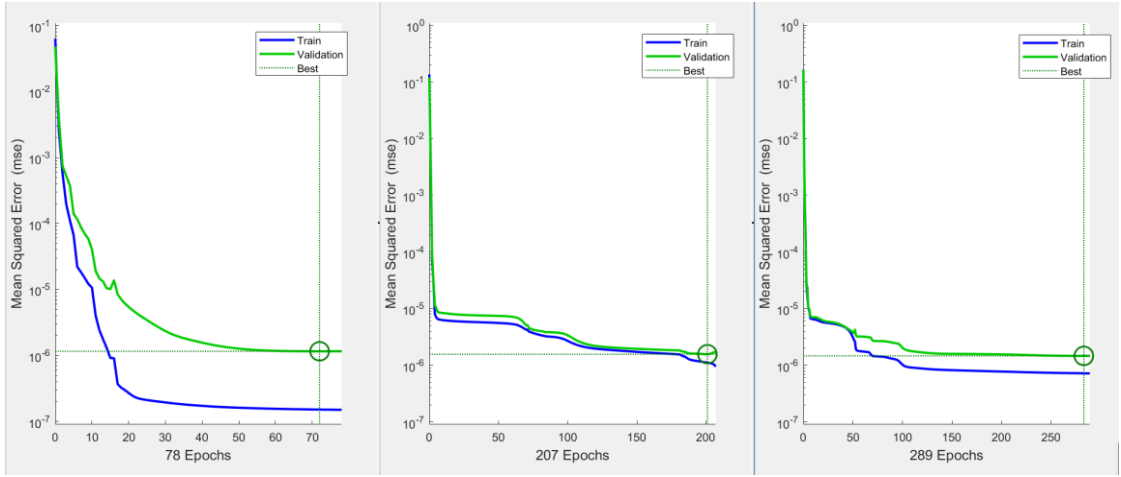

Figure 4. Training and Validation Loss on Lydia-Pinkham (left), Google (middle) and Facebook (right)

Figure 4 illustrates the training and validation loss of the training epochs. For the three DNN models, we can observe that the best validation performance with the lowest MSE occur at a certain epoch during the training process. Moreover, we use a linear regression to further investigate the DNN performance. The fitting curves between the predicted and observed market shares on the three datasets are shown in Figure 5, and Figure 6 presents histograms of the DNN prediction error. From Figures 5 and 6, each DNN model yields a strong predictive power and provides a good approximation.

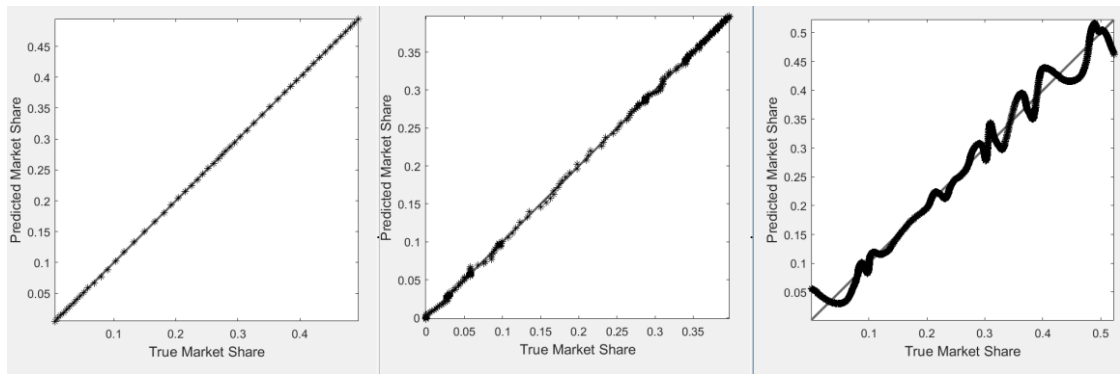

Figure 5. Predicted Market Shares on Lydia-Pinkham (left), Google (middle) and Facebook (right)

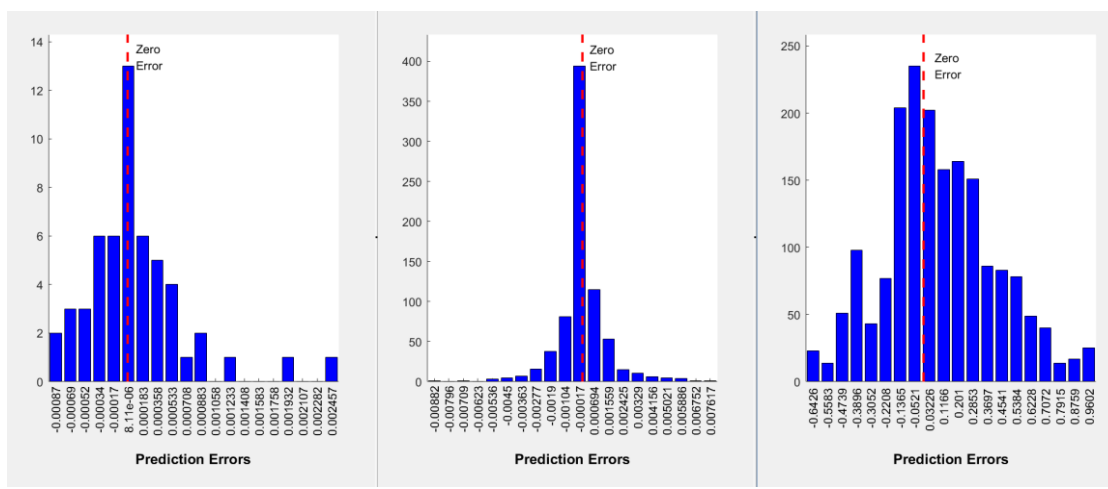

Figure 6. Histograms of Prediction Error on Lydia-Pinkham (left), Google (middle) and Facebook (right)

Based on the DNN trained models, the parameter estimates of the GVW model on the three datasets are given in Table 1.

Table 1. Parameter Estimates

| Datasets | Parameters | | | | MSE |
|---|---|---|---|---|---|
| | ρ | α | β | δ | |
| Lydia Pinkham | 4.037 E-05 | 0.801 | 0.675 | -0.010 | 2.820 E-06 |
| Google | 9.537 E-04 | 0.422 | 0.948 | -3.556 E-04 | 7.150 E-08 |
| Facebook | 1.435 E-04 | 0.688 | 0.590 | -4.117 E-05 | 7.180 E-07 |

From Table 1, we can see the following phenomena focusing on the ad elasticity index and the WoM index. (1) As for the ad elasticity index (α), its estimated value ranges from 0.422 to 0.801. In other words, the effect of diminishing returns is largest in search advertising, followed by social media advertising, and the traditional advertising exerts the least effect. The possible reason is that, compared to other advertising forms such as social media advertising, people involved in search advertising have more specific goals[23], thus advertising campaigns have less information effect; and the Lydia Pinkham advertising happened in the last century, at that time there was less competitive advertisers, thus advertising campaigns have more chances to obtain effectiveness.

(2) As for the WoM index ($\beta$), its estimated value ranges from 0.590 to 0.948. In other words, the effect of the WoM (i.e., $(1-\beta)$) is largest in social media advertising, followed by the traditional advertising, and search advertising exerts the least effect. The possible explanation for this phenomenon lies in the communication level characterized in advertising environments. That is, social media advertising facilitates online communications among people, which thus improve the WoM effect of advertising campaigns; in the traditional advertising, people communicate offline through face-to-face talking or other traditional manners, where the WoM effect works well; and in search advertising, people individually search for relevant information through search engines, thus has less chance to communicate between each other, which leads to the least WoM effect.

(3) In our case, the estimated value of the ad elasticity index is closed to 0.5 in search advertising ($\alpha = 0.422$) only, and that of the WoM index is closed to 0.5 in social media advertising ($\beta = 0.590$) only. Thus, we can conclude that, although the VW derivatives with $\alpha = 0.5$ and/or $\beta = 0.5$ make the analytical analysis possible, they are rarely realistic in practical situations. In this sense, this also confirms the advantage of the GVW model over the VW model and its derivatives.

### 6.3 Sensitivity Analysis

Next, we design computational experiments to conduct a sensitivity analysis with respect to modeling indexes. In particular, we focus on the ad elasticity and the WoM indexes because the two newly added indexes make the GVW model distinguishable from other VW-type models. For each model index, we investigate its effect on the relationships between market share and advertising expenditure. The datasets used in the following experiments were generated from historical advertising logs. In the following experiments, the ad effective index is set as 0.10, the decay index is set as 0.01.

First, we investigate the influence of the ad elasticity index (α). In this experiment, the WoM index is set as 1.0, for comparison purposes. Figure 7 illustrates market share at different advertising levels, with different ad elasticity indexes. Note that the curve with $\alpha = 1.0$ is defined by the VW model.

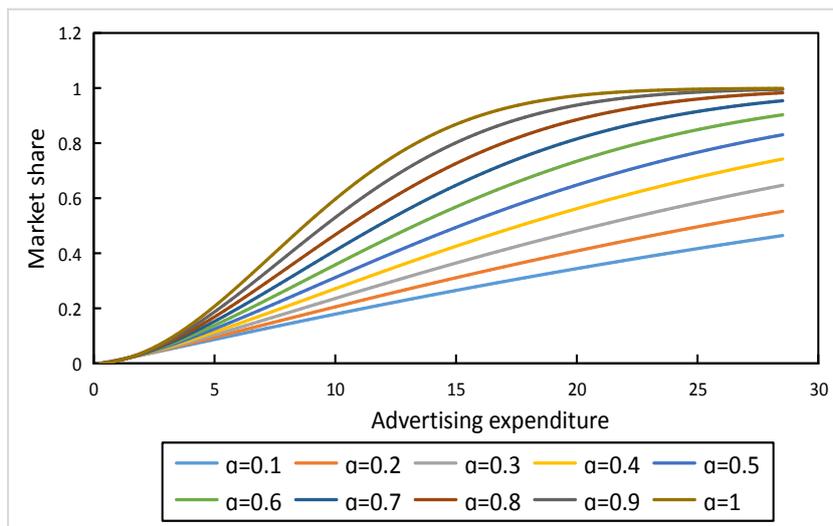

Figure 7. The Relationship between Market Share and Advertising Expenditure with Different Ad Elasticity Indexes

From Figure 7, we can observe the followings. Overall, given a certain advertising level, market share monotonically increases with the ad elasticity index. We also notice an interesting phenomenon. That is, the relationship between market share and advertising expenditure is concave in the situation with small ad elasticity indexes, which reveals decreasing marginal returns on advertising expenditure; however, it exhibits S-shape in the situation with large values of ad elasticity, which captures the phenomena of both increasing and decreasing marginal returns on various levels of advertising expenditure. In particular, the VW model with $\alpha = 1.0$ shows S-shape responses, while its derivative model with $\alpha = 0.5$ exhibits nearly concave responses.

Next, we study the influence of the WoM index (β). In this experiment, the ad elasticity index is set as 1.0, for comparison purposes. Figure 8 illustrates market share at different advertising levels, with different WoM indexes. Note that the curve with $\beta = 1.0$ is defined by the VW model.

From Figure 8, we can see that, given a certain advertising level, a larger WoM effect (i.e.,

a smaller β) results in a higher market share. This is because a smaller WoM index means a higher level of communications between the sold portion and the unsold portion of the potential market, which enlarges the advertising effectiveness from each unit of expenditure. Moreover, the advertising responses are S-shape regardless of the WoM effect. This indicates that the WoM index has little effect on the shape of advertising responses.

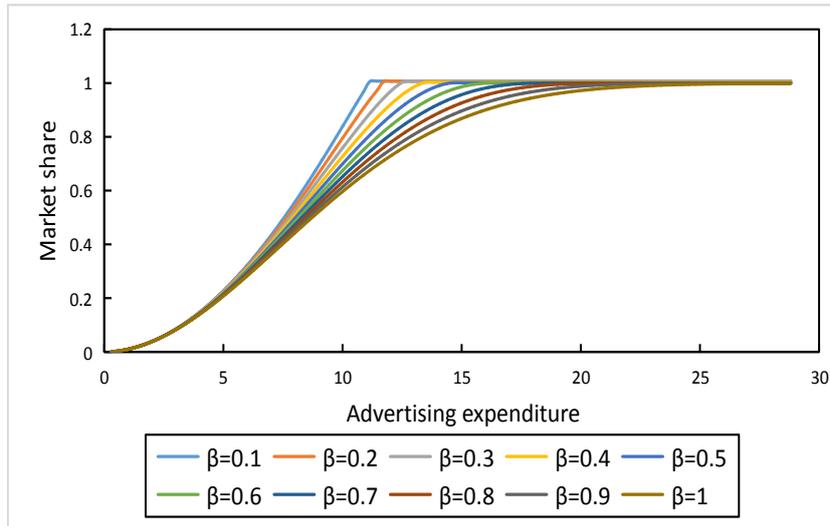

Figure 8. The Relationship between Market Share and Advertising Expenditure with Different WoM Indexes

### 6.4 Comparison with Econometric Models

In this section, we discuss potential advantages of VW-type models, especially the proposed GVW model, over econometric models of advertising.

First of all, compared to econometric models, VW-type models describe the relationship between market responses and advertising expenditure in a more parsimonious way[5]. Moreover, they are more suitable for optimal advertising decisions over time due to their differential-equation forms[2].

Next, we make a comparison between the GVW model and a commonly accepted form of econometric models with respect to several interesting phenomena of advertising effectiveness drawn from advertising practices[11]. As common in the literature, the econometric model of advertising for comparison takes sales in unit as the dependent variable ($s_t$), and advertising expenditure ($b_t$) and the lagged dependent variable ($s_{t-1}$) as independent variables, as is given by Equation (17). Note that sales can be conveniently transformed to market share by dividing a constant denoting the total sales, thus we use the two terms interchangeably in this comparison. Hereafter, we use Econbase as the short name for the econometric model defined by Equation (17).

$$\log s_t = \log c_0 + c_1 \log s_{t-1} + c_2 \log b_t + \mu_t \quad (17)$$

Through a systematic comparison between the GVW model and the Econbase model, we can obtain the following conclusions. Note we omit the details because of the lack of space. (1) Although the GVW model and the Econbase model show a similar pattern of market share over time, as a response to a rectangular pulse of advertising, there is some difference in the evolution after the cessation of advertising campaigns. (2) Both the GVW model and the Econbase model capture the steady-state response and the carryover effect of advertising in a similar manner. (3) Although both models capture the phenomenon of diminishing returns, the Econbase model accompanies a strict condition. (4) The GVW model takes into account the saturation level and the WoM effect, which are ignored by the Econbase model.

## 7. Conclusions and Future Work

In this research, we present a generalized form of VW model (GVW) that explicitly represents advertiser's elasticity with respect to advertising expenditure and the WoM effect among potential consumers by two additional indexes. Moreover, we discuss some desirable properties of the GVW model through analytical analysis, and present a deep neural network (DNN)-based estimation method to learn its parameters. Based on three realworld datasets, Computational experiments have been conducted to evaluate the GVW model and identified properties. Furthermore, we discuss potential advantages of VW-type models (especially the GVW model) over econometric models.

From the methodological perspective, the GVW model and its deep learning-based estimation method could support a rich set of big data-driven advertising analytics and decision makings for advertisers in various advertising forms.

This research generates several interesting insights. First of all, our empirical results highlight that it's against reality to fix the two newly added indexes, namely the ad elasticity index and/or the WoM index, as constants such as 1.0 and 0.5. This challenges the prior research in this stream, although such a treating brings a big convenience for analytical analysis.

Second, both the ad elasticity index and the WoM index have significant influences on the relationships between market share and advertising expenditure. On one hand, given a certain level of advertising expenditure, a higher ad elasticity ($\alpha$), or a higher WoM effect (i.e., $(1 - \beta)$) leads to a larger market share. Moreover, advertising responses exhibit concave when the ad elasticity is small, while they become S-shaped when the ad elasticity is large; the WoM index has little effect on the shape of advertising responses.

Third, the ad elasticity index and the WoM index affect the steady-state market share in different ways. The effect of the ad elasticity on the steady–state market share depends on whether the steady state market share is above or below a threshold value. In other words, there exists a critical steady–state market share. Above it the steady–state market share monotonically increases with the ad elasticity; below it the steady–state market share monotonically decreases with the ad elasticity. As for the WoM index, the steady–state market share monotonically decreases with it.

Fourth, the GVW model has potential advantages over econometric models of advertising, in terms of several interesting phenomena drawn from advertising practices, including the saturation level, the WoM effect and diminishing returns. Note that econometric models can capture diminishing returns on a strict complex condition related to multiple coefficients and advertising variables.

In this direction, several interesting perspectives deserve further research efforts: (1) estimation methods for time-varying parameters of the GVW model and empirical studies in different advertising media; (2) optimal advertising strategy based on the GVW; and (3) application of the GVW model to multi-channel advertising decisions due to its strength of encoding the potential heterogeneity among different media vehicles.

**Acknowledgment**

The authors are thankful to anonymous reviewers who provided valuable suggestions that led to a considerable improvement in the organization and presentation of this manuscript. This work was supported in part by the National Natural Science Foundation of China under Grant 71672067, Grant 71621002, and Grant 71810107003, and in part by Key Research Program of Chinese Academy of Sciences under Grant ZDRW-XH-2017-3.

**References**

[1] F. M., Feinberg, "On continuous-time optimal advertising under S-shaped response," *Management Science*, vol. 47, no. 11, pp. 1476-1487, 2001.

[2] Y., Yang, D., Zeng, Y., Yang, and J., Zhang, "Optimal budget allocation across search advertising markets," *INFORMS Journal on Computing*, vol. 27, no. 2, pp. 285-300, 2015.

[3] Y., Yang, Y.C., Yang, B.J., Jansen, and M., Lalmas, "Computational Advertising: A Paradigm Shift for Advertising and Marketing?," *IEEE Intelligent Systems*, vol. 32, no. 3, pp. 3-6, 2017.

[4] Y., Yang, X., Li, D., Zeng, and B. J., Jansen, "Aggregate Effects of Advertising Decisions: A Complex Systems Look at Search Engine Advertising via an Experimental Study," *Internet Research*, vol. 28, no. 4, pp. 1079-1102, 2018.

[5] K. R., Deal, "Optimizing advertising expenditures in a dynamic duopoly," *Operations Research*, vol. 27, no. 4, pp. 682-692, 1979.

[6] N. I., Bruce, B. P. S., Murthi, and R. C., Rao, "A dynamic model for digital advertising: The effects of creative format, message content, and targeting on engagement," *Journal of Marketing Research*, vol. 54, no. 2, pp. 202–218, 2017.

[7] J., Jansen, M., Zhang, K., Sobel, and A., Chowdury, "Twitter Power: Tweets as Electronic Word of Mouth," *Journal of the American Society for Information Science and Technology*, vol. 60, no. 11, pp. 2169-2188, 2009.

[8] Y., Liu, Y., Chen, R., Lusch, H., Chen, D., Zimbra, and S., Zeng, "User-generated content on social media: predicting market success with online word-of-mouth," *IEEE Intelligent Systems*, vol. 25, no. 1, pp. 75-78, 2010.

[9] M. L., Vidale, and H. B., Wolfe, "An operations-research study of sales response to advertising," *Operations Research*, vol. 5, no. 3, pp. 370-381, 1957.

[10] G. E., Kimball, "Some industrial applications of military operations research methods," *Operations Research*, vol. 5, no. 2, pp. 201-204, 1957.

[11] J. D., Little, "Aggregate advertising models: The state of the art," *Operations Research*, vol. 27, no. 4, pp. 629-667, 1979.

[12] H. I., Mesak, A., Bari, and Q., Lian, "Pulsation in a competitive model of advertising-firm's cost interaction," *European Journal of Operational Research*, vol. 246, no. 3, pp. 916-926, 2015.

[13] Y., Yang, J., Zhang, R., Qin, J., Li, F. Y., Wang, and W., Qi, "A budget optimization framework for search advertisements across markets," *IEEE Transactions on Systems, Man, and Cybernetics-Part A: Systems and Humans*, vol. 42, no. 5, pp. 1141-1151, 2012.

[14] G. M., Erickson, "Advertising competition in a dynamic oligopoly with multiple brands," *Operations Research*, vol. 57, no. 5, pp. 1106-1113, 2009.

[15] G. M., Erickson, "Note: Dynamic conjectural variations in a Lanchester oligopoly," *Management Science*, vol. 43, no. 11, pp. 1603-1608, 1997.

[16] S. P., Sethi, "Deterministic and stochastic optimization of a dynamic advertising model," *Optimal Control Applications and Methods*, vol. 4, no. 2, pp. 179-184, 1983.

[17] G., Sorger, "Competitive dynamic advertising: A modification of the Case game," *Journal of Economic Dynamics and Control*, vol. 13, no. 1, pp. 55-80, 1989.

[18] J. H., Case, *Economics and the competitive process*, New York University Press, 1979.

[19] N. Buduma, Fundamentals of Deep Learning, Designing Next-Generation Machine Intelligence Algorithms. O'Reilly Media, Sebastopol, CA, 2017.

[20] Y., LeCun, Y., Bengio, and G., Hinton, "Deep learning," *Nature*, vol. 521, no. 7553, pp. 436-444, 2015.

[21] H. Yu and B. M. Wilamowski, Levenberg–Marquardt Training, in: Industrial Electronics Handbook. vol. 5–Intelligent Systems, ed: CRC Press, pp. 12.1-12.15, 2011.

[22] R. W., Pollay, "Lydiametrics: Applications of econometrics to the history of advertising," *Journal of Advertising History*, no.1, pp. 3-18, 1979.

[23] Y. Yang, B. J. Jansen, Y. Yang, X. Guo, and D. Zeng, "Keyword optimization in sponsored search advertising: A multi-level computational framework," *IEEE Intell. Syst.*, vol. 34, no. 1, pp. 32–42, Jan./Feb. 2019.

**Short author bios**

Yanwu Yang is a full professor in the School of Management, Huazhong University of Science and Technology, and head of the ISEC research group (Internet Sciences and Economic Computing). His research interests include computational advertising, advertising decisions, web personalization, and user modeling. Yang has a PhD in computer science from the graduate school of ENSAM (École Nationale Supérieure d'Arts et Métiers). Contact him at yangyanwu.isec@gmail.com.

Baozhu Feng is currently a PhD candidate with the school of management, Huazhong University of Science and Technology. She is currently a researcher in the ISEC research group (Internet Sciences and Economic Computing). Her current research interests include online advertising. She is the corresponding author and can be contacted at fengbaozhu2019.isec@gmail.com

Daniel Zeng is Gentile Family Professor in the Department of Management Information Systems at the University of Arizona, and a Visiting Research Fellow at the Chinese Academy of Sciences. His research interests include intelligence and security informatics, infectious disease informatics, social computing, recommender systems, software agents, spatial-temporal data analysis, and business analytics. He has a PhD in industrial administration from Carnegie Mellon University. He has published one monograph and more than 330 peer-reviewed articles. He served as the editor-in-chief of IEEE Intelligent Systems from 2013-2016 and currently serves as President of the IEEE Intelligent Transportation Systems Society. Contact her at zeng@email.arizona.edu.